 \documentclass[letterpaper, 10 pt, conference]{ieeeconf}  

\IEEEoverridecommandlockouts                              

\usepackage{graphics} 
\usepackage{epsfig} 
\usepackage{mathptmx} 
\usepackage{times} 
\usepackage{amsmath} 
\usepackage{amssymb}  
\usepackage{bm}
\usepackage{algorithm}
\usepackage{multirow}
\usepackage{verbatim}
\usepackage{graphicx}
\usepackage{hyperref}
\usepackage[noend]{algpseudocode}
\usepackage{tikz}
\usepackage{booktabs} 
\usepackage{graphicx}
\usepackage{url}
\usepackage{pgfplots}
\usepackage{caption}
\usepgfplotslibrary{statistics}

\usepackage{subfigure}
\usepackage{color}
\usetikzlibrary{fit}
\usetikzlibrary{shapes,snakes}

\title{\LARGE \bf
   Group Surfing: A Pedestrian-Based Approach to Sidewalk Robot Navigation
}

\author{Yuqing Du$^{1}$, Nicholas J. Hetherington$^{1}$, Chu Lip Oon$^{1}$, Wesley P. Chan$^{1}$, Camilo Perez Quintero$^{1}$,\\ Elizabeth Croft$^{2}$, and H.F. Machiel Van der Loos$^{1}$ 
\thanks{$^{1}$Collaborative Advanced Robotics and Intelligent Systems (CARIS) Laboratory, University of British Columbia. $^{2}$Monash University
        }%
\thanks{© 2019 IEEE.  Personal use of this material is permitted.  Permission from IEEE must be obtained for all other uses, in any current or future media, including reprinting/republishing this material for advertising or promotional purposes, creating new collective works, for resale or redistribution to servers or lists, or reuse of any copyrighted component of this work in other works.}
}

\begin{document}
\maketitle
\thispagestyle{empty}
\pagestyle{empty}


\begin{abstract}
In this paper, we propose a novel navigation system for mobile robots in pedestrian-rich sidewalk environments. Sidewalks are unique in that the pedestrian-shared space has characteristics  of both roads and indoor spaces. Like vehicles on roads, pedestrian movement often manifests as linear flows in opposing directions. On the other hand, pedestrians also form crowds and can exhibit much more random movements than vehicles. Classical algorithms are insufficient for safe navigation around pedestrians and remaining on the sidewalk space. Thus, our approach takes advantage of natural human motion to allow a robot to adapt to sidewalk navigation in a safe and socially-compliant manner. We developed a \textit{group surfing} method which aims to imitate the optimal pedestrian group for bringing the robot closer to its goal. For pedestrian-sparse environments, we propose a sidewalk edge detection and following method. Underlying these two navigation methods, the collision avoidance scheme is human-aware. The integrated navigation stack is evaluated and demonstrated in simulation. A hardware demonstration is also presented.
%
%
%
%
%
%
%
%
%
%
%
%
%
%
\end{abstract}

\section{Introduction}
        \label{sec:introduction}

Robots are increasingly being introduced into human environments. To facilitate human-robot coexistence, robots must be capable of navigating these environments in a safe and socially-compliant manner. Such applications include mobile robots operating in elderly homes \cite{roy2003planning}; in hospitals \cite{5420402}; or in crowded public areas \cite{triebel2016spencer}. However, classical dynamic navigation algorithms are challenged by stochastic and human-populated environments. These deterministic algorithms are less viable when there is more uncertainty and when movements of humans and robots can affect each other \cite{trautman2015robot}. One common artifact of traditional navigation methods is the freezing robot problem (FRP) where dense crowds cause the robot to be unable to move due to the belief that all possible paths will lead to collisions \cite{5654369}, \cite{trautman2015robot}. Approaches that cause FRP do not account for the likelihood that people will adjust their path to allow passage for the robot (as they do for other pedestrians). 

As robot technology is used in more everyday applications, a crucial environment for human-robot coexistent navigation is pedestrian sidewalks. Consider the case where a mobile robot is used for delivering packages to homes. Such a robot must be capable of achieving two main goals: 1) safely sharing the sidewalk space, and 2) navigating to its goal as efficiently as possible. 

Sidewalks present a unique yet challenging environment in that the navigable space combines elements of both roads and free indoor spaces. Often sidewalk motion is restricted to two linear directions and the resulting navigable space is limited, like on roads. However, pedestrians generally do not walk in perfect queues. Instead, people tend to walk in groups of variable sizes and speeds \cite{laxman2010pedestrian} and move along with a general self-organizing crowd flow \cite{helbing2001self}. Compared to autonomous road navigation, sidewalk navigation must also account for stochastic human movement that necessitates dynamic obstacle avoidance. Furthermore, certain social rules, such as walking in lanes or affording more space in the direction of walking than in the perpendicular direction \cite{helbing2001self}, are rules that a robot should follow as well. 

\begin{figure}
\centering
\includegraphics[width=\linewidth] {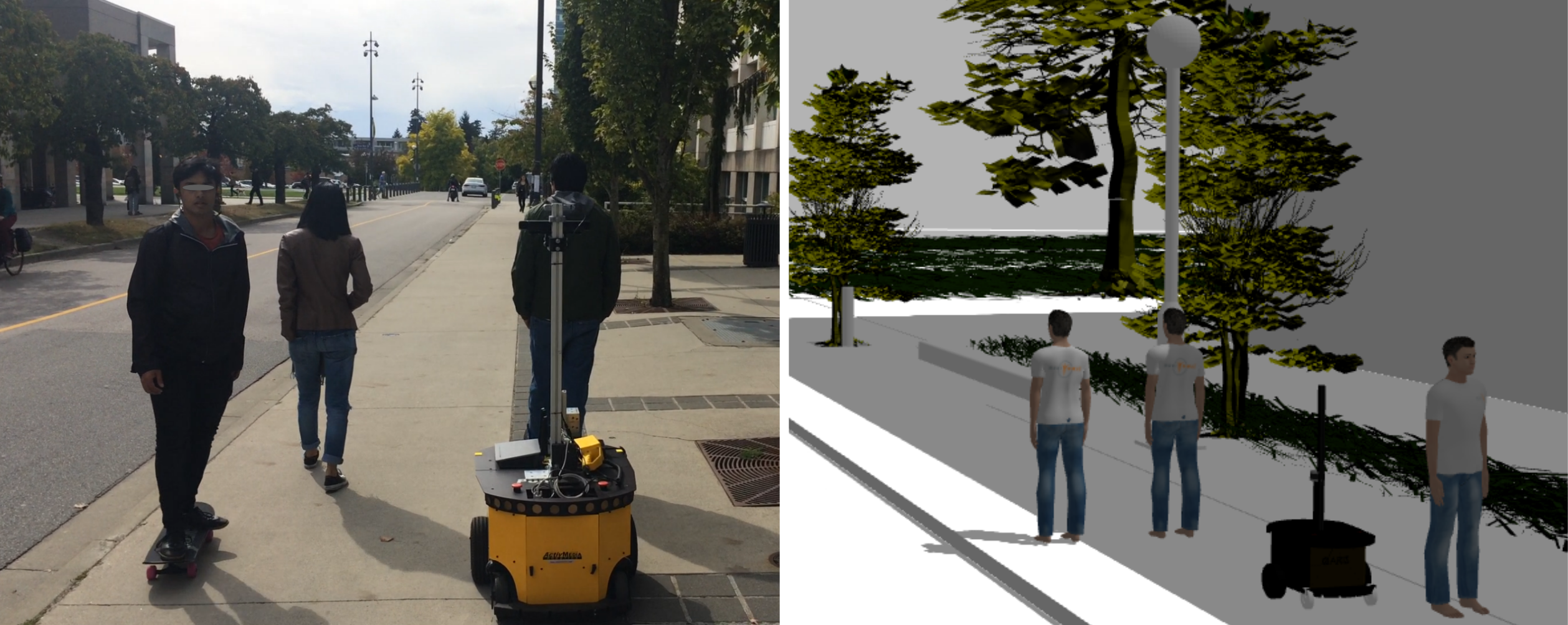}
\caption{Our mobile robot navigating the real and simulated environments using group surfing.}
\label{fig:sidewalk_navigation}
\end{figure}
	
Current research in human-aware path planning takes a variety of approaches. In general, methods are based on either reasoning, learning, or a combination thereof. Social rules, such as pedestrian conventions and appropriate proxemics, can be induced through either cost functions in reasoning approaches, or reward functions in reinforcement-learning based approaches \cite{Kruse13}. Examples include using inverse reinforcement learning to learn human-like navigation \cite{5509772}; people tracking combined with an iterative A* planner \cite{muller2008socially}; and modelling pedestrian intention to better predict movements of neighbouring people \cite{tamura2012development}. While each of these approaches take advantage of understanding human motion, they are often more suitable for open environments. In contrast, our approach specifically handles navigation in the more restrictive sidewalk setting. Here, the aforementioned approaches may be less effective as they do not account for the physical sidewalk boundaries, or how robot movement will affect pedestrian flow.

The key research question this paper considers is how mobile robots can utilize nearby pedestrian behaviours and flows to navigate towards a global goal. To explore this question, we have developed a system that imitates 
pedestrian behaviour. When our navigation stack detects people moving towards the robot's goal, a `group surfing' behaviour is used. This allows the robot to imitate and participate in pedestrian social behaviours. While methods for autonomous navigation in social environments have been proposed \cite{7759200}, our system is specialized for sidewalk navigation. In an unpopulated and simple sidewalk environment, the default behaviour is to follow a trajectory offset from the sidewalk curb towards the goal. Simultaneously, the underlying collision avoidance algorithm is capable of socially-aware behaviour to improve pedestrian comfort and safety in the shared space.
%
%

The organization of this paper is as follows. In Section \ref{sec:System Description} we propose the complete navigation system. In Section \ref{sec:Methods} we present the development of the three main navigation components that constitute the system. Section \ref{sec:Sim_demo_experiments} presents simulations for demonstrating and evaluating our approach. Section \ref{sec:HW_demo} describes our implementation in a real mobile robot and a demonstration of our navigation components. We provide some concluding remarks and directions for future research in Section \ref{sec:Conclusions}.

      
\section{System}
        \label{sec:System Description}



\begin{figure}[h]
\centering
\includegraphics[width=.5\textwidth] {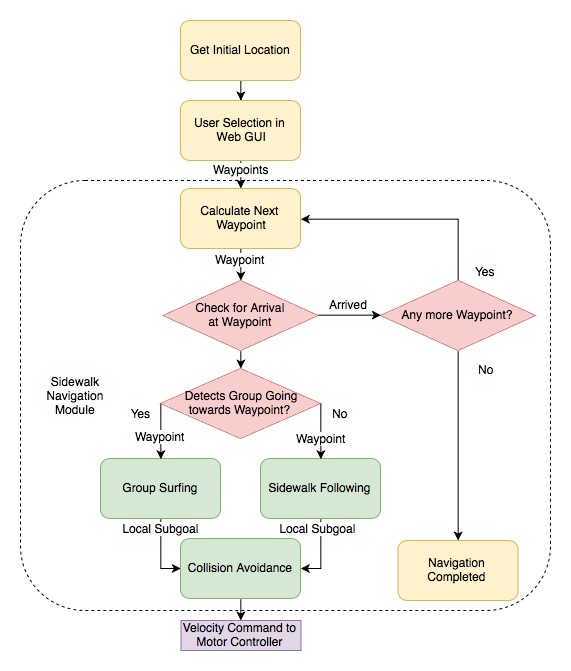}
\caption{Navigation Flow Diagram}
\label{fig:flow_diagram}
\end{figure}

We consider package delivery as our example task. A flow diagram of our system is shown in Figure ~\ref{fig:flow_diagram}. After the user selects a global goal $G$ (i.e., delivery location) using a graphical user interface (GUI), a path is computed by Google Maps' API with waypoints $W_n$ located at each intersection between the robot's initial location and $G$. The sidewalk navigation module is in charge of moving the robot towards the next waypoint. It uses one of two navigation modes depending on whether there is nearby pedestrian flow. If pedestrians moving towards a given waypoint are detected, the navigation module employs a `group surfing' method which relies on nearby pedestrian motion on the sidewalk for autonomous navigation. However, in the absence of humans, the robot will utilize a contextual `curb following' method. Each of these  methods generate subgoals for the robot to navigate to, and in either case, the subgoals are sent to a collision avoidance algorithm which produces a velocity command.

At each cycle, the system will also continuously check for the robot's arrival at the given waypoint. If the robot has reached the waypoint within a predefined tolerance, the module checks if there are more waypoints. If more waypoints exist, the current waypoint is updated to the next one. Otherwise, the module indicates that the robot has reached its goal and navigation is complete.

\section{Methods}
        \label{sec:Methods}
        \subsection{Group Surfing}\label{subsec:group_surf}
When given a waypoint to navigate to, `group surfing' is the preferred method of navigation. Our group surfing algorithm takes advantage of natural pedestrian behaviours through imitation. Such behaviours include: walking in lanes, avoiding collisions with other pedestrians or obstacles \cite{helbing2001self}, waiting at intersections to cross, and not walking into traffic. As a result, many of the challenges in collision avoidance and remaining on the sidewalk are mitigated. To imitate these movements, the algorithm constantly computes the preferred pedestrian group location and sets this as a navigation subgoal. 
%
%

Humans are detected through methods detailed in Subsection \ref{subsec:HW}. The detected people are then fed through a people tracking pipeline from the SPENCER project to identify tracked groups of individual(s) \cite{7487766}. Groups are classified using social relations, which the SPENCER project defines as ``spatial relations between people via coherent motion indicators".
\newline

\subsubsection{Filter Candidate Groups} 
At the lowest level, the robot should only imitate groups moving towards the waypoint. Consider the position and velocities of all parties in the global coordinate frame. Let the position of the waypoint (W) be $\bm{x}_{W}$ and the position of the robot be $\bm{x}_R$. The position of the waypoint relative to the robot is $\bm{x}_I = \bm{x}_{W} - \bm{x}_R$. 

The tracked groups are sets $G_i$, where $i$ is a positive integer value from the set $\{1, \ldots, n_G\}$ and $n_G$ is the total number of groups. Each set $G_i$ cannot be empty and contains pedestrians $p_j$ where $j$ is a positive integer value from the set $\{1, \ldots, n_{G_i}\}$ and $n_{G_i}$ is the total number of people in group $G_i$. From the velocities $\bm{v}_{p_j}$ and positions $\bm{x}_{p_j}$ of each pedestrian in a group $G_i$, we compute the average group velocity $\bm{v}_{G_i}$ and determine the $\bm{x}_{p_j}$ closest to $\bm{x}_R$, giving the closest person $p_{closest}$ for each group.

Given this information, we then filter out groups moving away from the waypoint. For each $G_i$, compute $\bm{v}_{G_i} \cdot \bm{x}_I$. If this value is non-positive, discard $G_i$ as a subgoal candidate. 

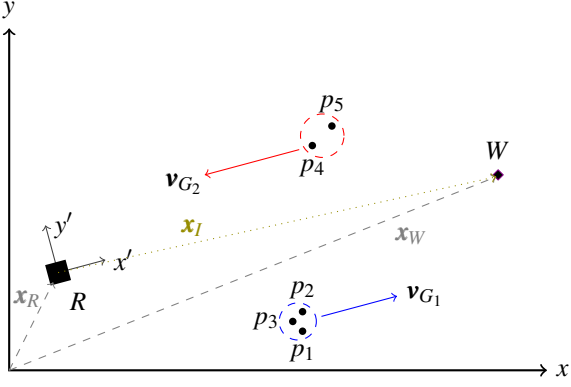
\begin{figure}[t]
\begin{tikzpicture}[scale=1.3]
    \draw [<->,thick] (0,3.5) node (yaxis) [above] {$y$}
        |- (5.5,0) node (xaxis) [right] {$x$};
    \begin{scope}[rotate around={15:(.5,1)},draw=darkgray]
        \draw[->] (.5,1) -- (1,1)  node[right, text width=5em] {$x'$};
        \draw[->] (.5,1) -- (.5,1.5)  node[right,text width=5em] {$y'$};
    \end{scope}
    \draw (.5,1) node[rectangle,fill,inner sep=4pt,label=below:$R$, rotate=15](r){};
    
    \draw[->, color=gray, dashed] (0,0) -- (.45,.9) node[label={[label distance=-1cm]35:$\bm{x}_{R}$},text width=5em]{};
    
    \begin{scope}[rotate around={15:(3,.5)},draw=blue]
        \draw[->] (3.2,.5) -- (4,.5)  node[right, text width=5em] {$\bm{v}_{G_1}$};
    \end{scope}
    
    \draw (3, .4) node[circle, fill, inner sep=1pt, label=below:$p_1$](p_1){};
    \draw (3, .6) node[circle, fill, inner sep=1pt, label=above:$p_2$](p_2){};
    \draw (2.9, .5) node[circle, fill, inner sep=1pt, label=left:$p_3$](p_3){};
    
    \node [circle, draw=blue, dashed,inner sep=0.5pt, fit=(p_1) (p_2) (p_3)] {};
    
    \begin{scope}[rotate around={15:(2,2)},draw=red]
        \draw[->] (3,2) -- (2,2)  node[label={[label distance=-1cm]30:$\bm{v}_{G_2}$},text width=5em] {};
    \end{scope}
    
    \draw (3.1, 2.3) node[circle, fill, inner sep=1pt, label=below:$p_4$](p_4){};
    \draw (3.3, 2.5) node[circle, fill, inner sep=1pt, label=above:$p_5$](p_5){};
    
    \node [circle, draw=red, dashed,inner sep=0.5pt, fit=(p_4) (p_5)] {};
    
    \draw (5, 2) node[diamond, draw=violet, fill, inner sep=1pt, label=above:$W$](W){};
    \draw[->, color=gray, dashed] (0,0) -- (4.98,1.98) node[label={[label distance=-2cm]35:$\bm{x}_{W}$},text width=5em]{};

    \draw[->, color=olive, dotted] (.5,1) -- (4.98, 1.98) node[label={[label distance=-5cm]12:$\bm{x}_{I}$},text width=5em]{};

\end{tikzpicture}
\caption{In the global coordinate frame, we compute whether groups $G_i$ are suitable for following. Given each pedestrian group's average velocity $\bm{v}_{G_i}$ and the position of the waypoint relative to the robot $\bm{x}_I$, computing if $\bm{v}_{G_i} \cdot \bm{x}_I > 0$ determines whether $G_i$ will bring the robot closer to the waypoint. Here, $G_2$ is not a viable candidate as it is heading further from the waypoint.The $x'$ and $y'$ axes are in the robot's frame of reference and show the orientation of the robot.}
\label{fig:velocity_filter}
\end{figure}

\subsubsection{Smart Group Selection} 
Once we have filtered out unsuitable groups, the algorithm selects the optimal group to follow. We define the optimal group as follows. Given a desired speed, often the robot's maximum safe speed $v_{max}$, a group $G_n$ is better than a group $G_m$ if $v_{max}  - |\bm{v}_{G_n}|$ is a smaller positive value than $v_{max}  -  |\bm{v}_{G_m}|$. The robot will not follow a group where $|\bm{v}_{G_i}| > v_{max}$ as it will not be able to keep up with $G_i$ for more than a short moment in time. After determining the optimal group, the algorithm outputs the position of the $p_{closest}$ of that group as the group surfing subgoal. We intentionally select the closest person as a subgoal as attempting to reach the average group position could lead to path planning through pedestrians located between the average group position and the robot's current position. This subgoal selection is based on the assumption that $p_{closest}$ will move to a new location before the robot arrives. However, if this is not the case, the collision avoidance component will prevent the robot from colliding with $p_{closest}$. The optimal group at any given time is constantly updated, effectively causing the robot to `surf' between different pedestrian groups to maximize efficiency.

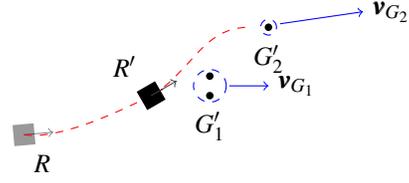
\begin{figure}[h]
\centering
\begin{tikzpicture}[scale=1.3]
   
    \begin{scope}[rotate around={5:(.5,1)},draw=darkgray, opacity=.6]
        \draw[->] (.5,1) -- (.8,1)  node[right, text width=5em] {};
    \end{scope}
    \draw (.5,1) node[rectangle,fill,inner sep=4pt,label=below:$R$, rotate=5, opacity=0.4](r){};
    
    \begin{scope}[rotate around={0:(2.6,1.5)},draw=blue]
        \draw[->] (2.6,1.5) -- (3,1.5)  node[right, text width=5em] {$\bm{v}_{G_1}$};
    \end{scope}
    
    \draw (2.4, 1.4) node[circle, fill, inner sep=1pt](p_1){};
    \draw (2.4, 1.6) node[circle, fill, inner sep=1pt](p_2){};
    
    \node [circle, draw=blue, dashed,inner sep=0.5pt, fit=(p_1) (p_2), label=below:$G_1'$] {};

    \begin{scope}[rotate around={15:(3,2)},draw=blue]
        \draw[->] (3.15,2.1) -- (4,2)  node[right, text width=5em] {$\bm{v}_{G_2}$};
    \end{scope}
    
    \draw (3, 2.1) node[circle, fill, inner sep=1pt](p_3){};
    
    \node [circle, draw=blue, dashed,inner sep=0.5pt, fit=(p_3), label=below:$G_2'$] {};
    
    \begin{scope}[draw=red, dashed]
    \draw    (1.8, 1.4) to[out=30,in=180] (2.8,2.1) {};
    \draw    (.5, 1) to[out=-10,in=30] (1.8, 1.4) {};
    
    \end{scope}

    \draw (1.8, 1.4) node[rectangle,fill,inner sep=4pt,label=above:$R'$, rotate=30](r){};
    \begin{scope}[rotate around={30:(1.8, 1.4)},draw=darkgray]
        \draw[->] (1.8, 1.4) -- (2.1,1.4)  node[above, text width=5em] {};
    \end{scope}

\end{tikzpicture}
\caption{Illustrating group surfing behaviour. $R$ is the robot's initial position and the dotted red line illustrates the path taken. Initially the algorithm will lead the robot to imitate the closer $G_1$ until it reaches $R'$, when $G_2'$ is within the field of view. Since $\bm{v}_{G_2} > \bm{v}_{G_1}$, the algorithm will switch to imitating $G_2'$. }
\label{fig:surf}
\end{figure}

In summary, given a typical sidewalk where there may be groups of pedestrians moving either towards or away from a waypoint, the algorithm will constantly output the position of the optimal group to imitate (if it exists) as a local subgoal.  

\subsection{Curb Following}\label{subsec:ped_ind}
\begin{figure}[ht]
\centering   
\subfigure[Robot following a group of people]{\label{fig:ca_a}
\includegraphics[trim={0 1cm 0 1cm},clip,width=.40\textwidth]
{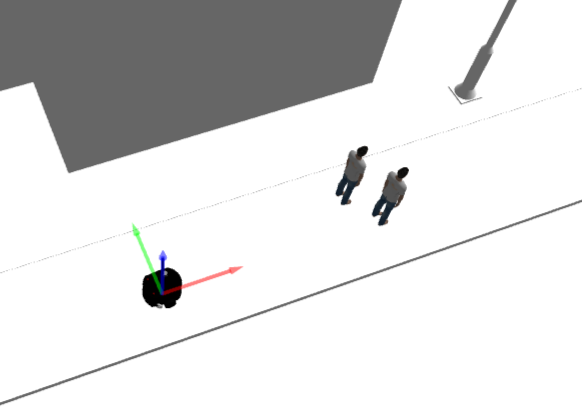}}
\subfigure[Visualization of people group, curb and wall detection and avoidance for the scenario in (a)]{\label{fig:ca_b}
\includegraphics[trim={0 1cm 0 1cm},clip,width=.40\textwidth]
{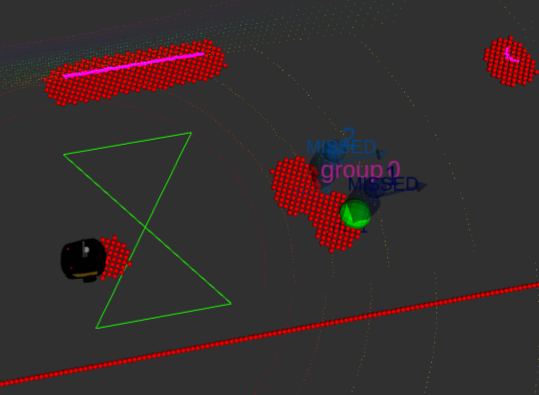}}
\caption{Group following and static obstacle avoidance. a) The robot following a group of pedestrians to its goal at the next intersection. b) Visualization showing: the curb detection described in Section \ref{subsec:ped_ind} (green polygon); static obstacle avoidance described in Section \ref{subsec:CA} (small red spheres); and the pedestrian detection and group goal (green sphere) described in Section \ref{subsec:group_surf}}
\end{figure}

Our system defaults to group surfing for navigation. However, when pedestrians are not in the field of view of the robot, a sidewalk detection navigation mode is activated to allow the robot to follow the curb. We make use of contextual knowledge; sidewalks are normally surrounded by streets and buildings or empty space. To find the boundary between the sidewalk and the street, our robot first acquires a surrounding point cloud using a 3D laser sensor and filters out points that are at the same height as or above the plane defined by the robot wheel contacts, resulting in a point cloud set $S$. Then the Random Sample Consensus (RANSAC) method~\cite{fischler1981random} is used to fit a plane on the remaining points by:
\begin{enumerate}
\item Selecting three non-collinear points in the filtered set $S$ and computing a candidate model.
\item  Computing distances from all other points in $S$ to the plane.
\item  Calculating a score value by counting the number of points inside of a 5 cm threshold. Steps one to three are performed for k iterations, the model with more inliers is selected.
\item  Using the inlier points in the winner set to calculate the model coefficients in a least-squares formulation. 
\item After projecting all the inlier points into the calculated plane, estimating a concave hull point cloud $C$ with $\alpha=5.0$~\cite{asaeedi2017alpha}, where $\alpha$ describes the smoothness level of the computed hull. 
\end{enumerate}  
The set $C$ is arranged in a kd-tree data structure and a nearest search algorithm is used to find the k nearest points to the robot. These points are used to fit a line that represents the curb (bottom red line in Figure~\ref{fig:ca_b}). Then, a subgoal point is set in the line that is parallel to the curb and passes through the robots' location.



\subsection{Collision Avoidance} \label{subsec:CA}
\subsubsection{Human-Aware Collision Avoidance}
Both while following the selected group or curb, we also avoid groups and individual pedestrians we are not following. We use a learned approach, Socially-Aware Collision Avoidance with Deep Reinforcement Learning (SA-CADRL) \cite{Chen17}, as the collision avoidance component of our navigation stack. The collision avoidance system navigates to a local subgoal generated by either the group surfing  or the curb following approach.

The SA-CADRL policy $\pi : \bm{s_t} \rightarrow \bm{a_t}$ maps a state $\bm{s_t}$ to an action $\bm{a_t}$. A state $\bm{s_t} = (P_{r_t},v_{r_t},P_{0_t},v_{0_t},...,P_{n_t},v_{n_t})$ is the robot's own state and its observation of surrounding pedestrians, namely their pose $P$ and velocity $v$. The corresponding action $\bm{a_t} = (\nu_t, \omega_t)$ is a control command consisting of a linear velocity $\nu_t$ and an angular velocity $\omega_t$,  which minimizes collisions and the time to goal and maximizes the social awareness of the robot's motion.

The reinforcement training process induces social awareness through social reward functions, which give higher values to actions that follow social rules. Maximal social awareness is considered to be achieved when an action follows two socially-aware behaviours: staying to the right; and passing on the left. This training used simulated data, but the policy was demonstrated with hardware in a populated shopping-mall-like environment. We did not train the policy ourselves, but used the results made publicly available by the authors of  \cite{cadrlROS}.


\subsubsection{Static Obstacle Avoidance}
In addition to following groups, avoiding individuals, and/or following the curb, we avoid static obstacles. These obstacles include street posts, bins, benches, walls, and curbs. We also use SA-CADRL to avoid these static obstacles by adding ``static pedestrians" to the state vector $s_t$. Using data from the curb detection detailed in Subsection \ref{subsec:ped_ind} and a 2D laser scan, we populate $s_t$ with small, stationary ``pedestrians" at the locations of the static obstacles. The action generated by the policy then avoids both these static obstacles and the real pedestrians.

\section{Simulation Demonstration and Experiments}
        \label{sec:Sim_demo_experiments}
        For validating the system in simulation, we built a simulation world modelled after real sidewalks, buildings, and intersections. We use the Robot Operating System (ROS) and Gazebo simulator suite. To simulate pedestrians, we use the Pedsim ROS \cite{pedsimROS} library, which relies on the social force model \cite{helbing1995social}. These pedestrians follow multiple paths in both directions down each sidewalk and may also cross intersections.

\subsection{Simulation Demonstration}\label{subsec:sim_demo}
For simulation testing and demonstrations, We created a replica of our mobile robot (described in Subsection \ref{subsec:HW}) and our research lab neighborhood. We calibrated the simulation with the real world using GPS coordinates, such that landmarks in the simulated world correspond to those in the real world. We then populated the world with simulated pedestrians. As a case study, we sent the robot from our lab to a coffee shop. A user selects a destination through a browser interface. Depending on the pedestrian flow and the robot's surroundings, the robot navigates by surfing across groups of people or by detecting the curb and following it. A detailed demonstration of our system is included in the attached video~\footnote{\url{https://youtu.be/fP33UHO_978}}.

\subsection{Simulation Experiments and Evaluation}
In evaluating our navigation system, our main goal was to show that the system successfully navigates the robot to its final goal through a socially-acceptable path. That is, the path that our robot takes to the goal is similar to what a pedestrian would take to the same goal.

To illustrate this quantitatively, we used the same simulation environment described in Subsection \ref{subsec:sim_demo}. In addition to the pedestrians shown in the demonstration video, we further populated the world with simulated pedestrians moving back and forth on the sidewalk at different velocities. The pedestrians were programmed to move forward on the right side of the sidewalk as is the norm. In this experiment, the robot's goal was to move to the same coffee shop as seen in the demonstration video. This path involves navigation down straight paths, left and right turns, and intersection crossing using both group surfing and curb following. 

The experiments we conducted are based on \cite{5509772} to evaluate whether the path taken by our robot is similar to that of a human in the same environment. Over 10 trials, we tracked the path taken by the robot ($r_{\textnormal{path}}$) and the path taken by a simulated pedestrian ($p_{\textnormal{path}}$). We also tracked the shortest path that the robot could take within the confines of the sidewalk ($s_{\textnormal{path}}$).

\begin{figure}
\label{fig:a}\includegraphics[width=.52\textwidth]{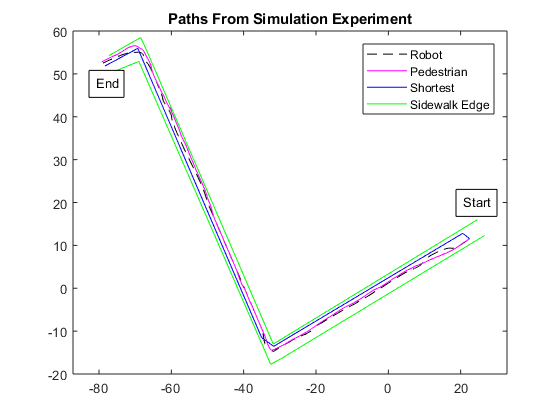}
\caption{Plotting paths taken by the robot, a pedestrian, and the most efficient path. The grid is in meters. Note that in the middle segment, the path of the pedestrian and robot both remain on the right side of the sidewalk while the shortest path requires moving on the left side, going against the norm.}
\end{figure}

To evaluate the similarity of the paths, we compared each trial of $p_{\textnormal{path}}$ to each trial of $r_{\textnormal{path}}$ and $s_{\textnormal{path}}$. This comparison was done by computing the directional Hausdorff distance from $p_{\textnormal{path}}$ to either $r_{\textnormal{path}}$ or $s_{\textnormal{path}}$ (collectively referenced as $p_2$ below), which is defined as

$$h_{\textnormal{directional}} = max_{\textbf{x} \in p_{\textnormal{path}}} \{ min_{\textbf{y} \in p_2} \{ d(\textbf{x},\textbf{y})\}\}$$

where $\textbf{x},\textbf{y}$ are coordinate points in $p_{\textnormal{path}}$ and $p_2$ respectively. The distance function $d(\textbf{x},\textbf{y})$ is defined as the Euclidean distance between ${\textbf{x}}$ and ${\textbf{y}}$. This is a measure of the maximum difference between two paths.

We also compute the Hausdorff `average' distance from $p_{\textnormal{path}}$ to $p_2$, defined as

$$ h_{\textnormal{average}} = \frac{\sum_{\textbf{x} \in p_{\textnormal{path}}} min_{\textbf{y} \in p_2} \{d(\textbf{x},\textbf{y})\}}{N_{p_{\textnormal{path}}}}$$ 

where $N_{p_{\textnormal{path}}}$ is the number of points in the pedestrian path. This is a measure of the average difference between two paths.

We conducted an independent samples t-test to test for differences in $h_{\textnormal{directional}}$ between the pedestrian and the robot (P-R $h_{\textnormal{directional}}$) and $h_{\textnormal{directional}}$ between the pedestrian and the shortest path (P-S $h_{\textnormal{directional}}$). There was a significant difference between P-R $h_{\textnormal{directional}}$ (M=1.97 m,  SD=0.07 m) and P-S $h_{\textnormal{directional}}$ (M=2.36 m, SD=0.07 m); p<.05. There was also a significant difference between P-R $h_{\textnormal{average}}$ (M=0.47 m, SD= 0.01 m) and P-S $h_{\textnormal{average}}$ (M=1.22 m, SD=0.04 M); p<.05. This means the robot followed an equivalent path to a test pedestrian.

From comparing each $p_{\textnormal{path}}$ trial to each $r_{\textnormal{path}}$ trial and the $s_{\textnormal{path}}$, we computed 100 values for $h_{\textnormal{directional}}$ and $h_{\textnormal{average}}$. The average values can be found in Figure \ref{fig:sim_exp_results}. The lower values of both $h_{\textnormal{directional}}$ and $h_{\textnormal{average}}$ for $r_{\textnormal{path}}$ than $s_{\textnormal{path}}$ suggest that the path taken by the robot is more similar to the path taken by a simulated pedestrian under the same conditions. Thus, we conclude that our navigation strategy is able to successfully navigate to the goal in a manner that is more socially acceptable than the shortest possible path.

\begin{figure}
\centering
\begin{tabular}{|l|l|l|}
\hline
 &  $h_{\textnormal{directional}}$ &  $h_{\textnormal{average}}$ \\ \hline
$r_{\textnormal{path}}$ comparison & 1.9661  &  0.4726 \\ \hline
$s_{\textnormal{path}}$ comparison & 2.3606& 1.2195 \\ \hline
\end{tabular}
\caption{Averaged $h_{\textnormal{directional}}$ and $h_{\textnormal{average}}$ across 10 simulated runs of pedestrians, the robot with our navigation strategy, and the shortest path. All values are in meters. }
\label{fig:sim_exp_results}
\end{figure}

\pgfplotsset{compat=1.10}
\pgfplotsset{
		box plot width/.initial=.4,
        boxplot/every box/.style={draw=black},
        boxplot/every whisker/.style={black},
    	boxplot/every median/.style={black}
}

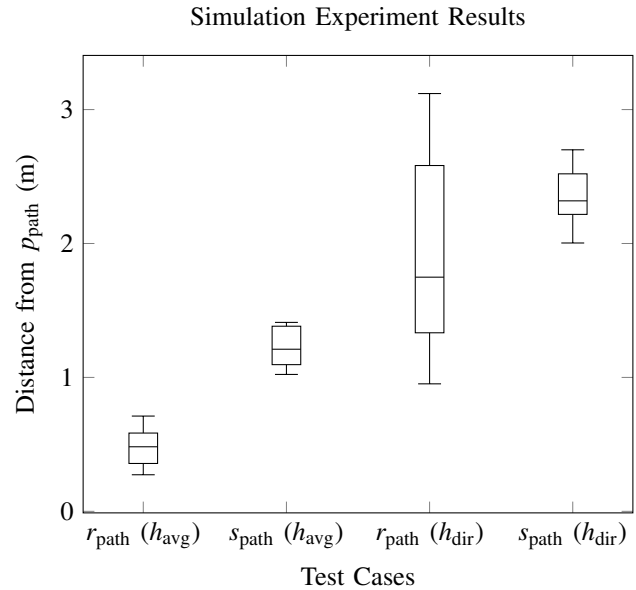
\begin{figure}
\begin{tikzpicture}
  \begin{axis}
    [
	boxplot/draw direction=y,
    width=0.5\textwidth,
    xtick={1,2,3,4},
    xticklabels={$r_{\text{path}}$ ($h_{\text{avg}}$),$s_{\text{path}}$ ($h_{\text{avg}}$),$r_{\text{path}}$ ($h_{\text{dir}}$),$s_{\text{path}}$ ($h_{\text{dir}}$)},
    xlabel=Test Cases,
   ylabel=Distance from $p_{\text{path}}$ (m),
   title=Simulation Experiment Results
    ]
    \addplot+[
    boxplot prepared={
      box extend=.2,
      median=0.482105405,
      upper quartile=0.584472294,
      lower quartile=0.357427304,
      upper whisker=0.711344146,
      lower whisker=0.273125036
    },
    ] coordinates {};
    \addplot+[
    boxplot prepared={
      box extend=.2,
      median=1.21049392,
      upper quartile=1.38226445,
      lower quartile=1.095358202,
      upper whisker=1.410839172,
      lower whisker=1.022249957
    },
    ] coordinates {};
    \addplot+[
    boxplot prepared={
      box extend=.2,
      median=1.748531524,
      upper quartile=2.582230554,
      lower quartile=1.333223323,
      upper whisker=3.119646574,
      lower whisker=0.952008027
    },
    ] coordinates {};
    \addplot+[
    boxplot prepared={
      box extend=.2,
      median=2.318449863,
      upper quartile=2.520150382,
      lower quartile=2.21734289,
      upper whisker=2.699586111,
      lower whisker=2.003750926
    },
    ] coordinates {};
    
  \end{axis}
\end{tikzpicture}
\caption{Box plot for simulation result data. For $h_{\text{avg}}$, the $r_{\text{path}}$ case is distinctly smaller than $s_{\text{path}}$ case across the dataset. For $h_{\text{dir}}$, the $r_{\text{path}}$ case has a smaller median than $s_{\text{path}}$ but a larger spread. We note that this likely resulted from the robot occasionally making wider turns than the simulated pedestrian, so the maximum difference between $r_{\text{path}}$ and $p_{\text{path}}$ can be large but the average difference is small. }
\end{figure}

\section{Hardware Demonstration}
         \label{sec:HW_demo}
         \begin{figure*}
\centering
{   \includegraphics[trim = 0mm 0mm 0mm 0mm, clip=true,width=.95\textwidth] 
{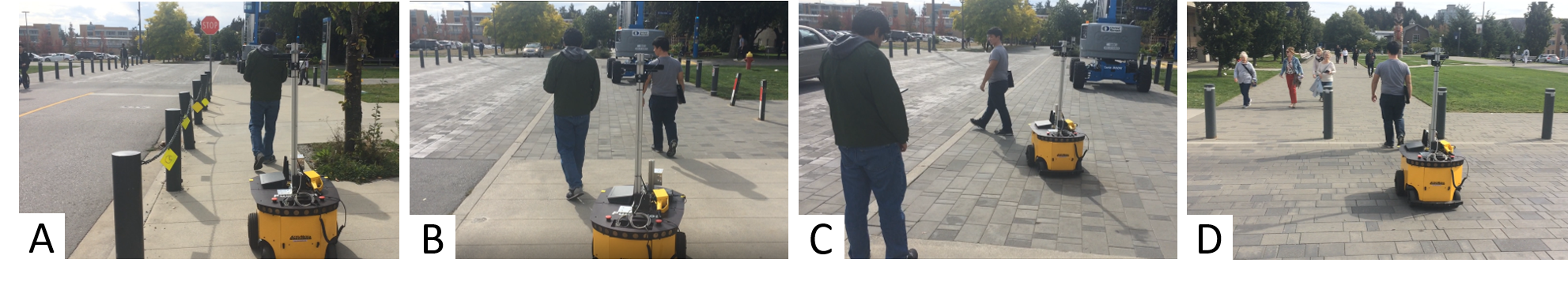} 
}
\caption{Group surfing navigation using the PowerBot  (A) The robot follows a pedestrian on the sidewalk (B) A second pedestrian with higher velocity appears in the robot visual field (C,D) The robot follows the second pedestrian crossing the road. 
}
\label{fig:real_robot_group_surfing}
\end{figure*}


\subsection{Hardware Setup}\label{subsec:HW}
We use the PowerBot from Omron Adept Mobile Robots 
as our differential drive mobile base.
The robot is equipped with multiple sensors: a Velodyne VLP-16 3D LiDAR sensor
; a SICK LMS-200 2D laser sensor
; a RealSense RGB-D sensor
, and GPS and IMU sensors. These sensors are both simulated in Gazebo and equipped on the real robot.
In simulation we use the Velodyne to track both the curb and pedestrians. However, in physical testing the Velodyne is only used for curb detection and the RGB-D sensor is used for pedestrian tracking.
To track people with the Velodyne in simulation, we compress its pointcloud into a 2D scan. This scan is input to an open source leg tracker \cite{legDetROS} 
whose output is then fed through the people tracking pipeline from the SPENCER project \cite{7487766}, \cite{SPENCERROS}. 
In our simulated testing we have found that occasionally round street poles are also detected as legs through this method. To reduce these false positives, we switched to an RGB-D-based upper body detector provided in the SPENCER project for testing within the real robot. The output of the RGB-D sensor is fed through the same people tracking pipeline from the SPENCER project. Group tracking results from SPENCER are used for our group surfing algorithm. The SPENCER group tracker models the evolution of groups through time such as forming, splitting, and merging by considering social relations among people \cite{linder2014}. 

\subsection{Demonstration and Discussion}

As a proof of concept, we implemented our group surfing method in the PowerBot and tested on the same route that the simulation was based on, see Figure~\ref{fig:real_robot_group_surfing}.
During our demonstration we let the robot navigate autonomously on a sidewalk and cross a road using our group surfing method. Some practical challenges that we encountered were: 

1) The system behavior relies on people detection, so we had to tune some of the RealSense camera parameters depending on the time of the day for reliable detection.    

2) The GPS waypoints calculated by the Google Maps' API are normally set in the middle of the road. Since we need to detect the waypoint from the sidewalk, we increased the waypoint threshold. Although our approach takes into consideration this GPS error, the threshold has to be modified accordingly for wider roads. Otherwise, the next waypoint will not be updated (see Section~\ref{sec:System Description}). 

3) Human walking speed varies depending on many factors. An experimental study~\cite{carey2005establishing} on young adults report that the average walking pedestrian speed is 1.47m/s. However, our PowerBot's max speed is 0.8m/s. This limits its capacity of following faster pedestrian groups. Thus, we asked pedestrians to walk at lower speeds during our tests.  

We also tested our curb following approach with no pedestrians on the sidewalk. In contrast with the simulated world, real-world sidewalks are not perfectly flat and streets have an elevation for drainage purposes. To account for this, we tuned the algorithm for this test scenario.  A video~\footnote{\url{https://youtu.be/PHhtYIkdxiw}} demonstrating group surfing and curb following is included.

\section{Conclusions}
        \label{sec:Conclusions}
In this paper, we have presented a novel system for navigation on sidewalks. Our group surfing method is a new navigation strategy based on the imitation of pedestrian social behaviour and observations of pedestrian flow behaviour. However, not all sidewalks are constantly populated. Thus, for pedestrian-independent navigation, we follow the curb. Simultaneously, we avoid static obstacles and other pedestrians with a socially-aware collision avoidance strategy. We demonstrated our system in simulation and evaluated our approach by comparing the path taken by the robot with that of a simulated pedestrian. Our experimental results suggest that our approach produces motion that is more socially compliant for the sidewalk environment. 
%
%

Future work will further develop and test our navigation system. For the group surfing component, one main area for improvement is in the selection process of groups to imitate. In particular, it would be beneficial to filter based on criteria in addition to group velocity, such as group trajectory or group size. With the current selection criteria, the robot may oscillate between groups that have very similar velocities and an improved algorithm should prevent such cases. Furthermore, external observers of the group surfing behaviour will be interviewed to gauge if the imitation behaviour is socially acceptable. For collision avoidance, a more specialized technique would allow for more efficient navigation. Future work to this end will involve a different reinforcement learning approach with data from a real sidewalk. Furthermore, we hope to decouple static collision avoidance from dynamic collision avoidance. The SA-CADRL policy also prioritizes constant motion, which results in arcing motions that are not suitable for the static constrictions of a sidewalk environment. For curb following, our approach only works for sidewalks that limit directly to the street, ignoring common tree belt, median, hellstrip, etc. Our future plan is to introduce detection and recognition of these non-transitable areas and incorporate them in our navigation module. Overall, more testing with different group sizes and sidewalk types should be conducted to support the generalizability of our system to sidewalk environments.
        

\section{Acknowledgements}
	\label{sec:Acknowledgements}
    This work was supported by the Natural Sciences and Engineering Research Council of Canada (NSERC) and Mitacs Canada. We thank Zhenyu Guo for his support, and 
the authors of \cite{Chen17} for making their trained SA-CADRL policy publicly available.










\bibliographystyle{IEEEtran}
\bibliography{bibliography}

\end{document}